\pdfoutput=1
\documentclass{article}

\usepackage[version=4]{mhchem}

\usepackage{hyperref}
\usepackage{url}
\usepackage{graphicx}
\usepackage{booktabs}
\usepackage{enumitem}
\usepackage{amsmath}
\usepackage{wrapfig}
\usepackage{mathtools}
\usepackage{float}
\usepackage{color, colortbl}
\usepackage{amsthm}
\usepackage{thmtools}
\usepackage{thm-restate}
\usepackage{epstopdf}
\usepackage{bbm}
\usepackage{dsfont}
\usepackage{threeparttable}
\usepackage{wrapfig,lipsum,booktabs}
\usepackage[capitalize,noabbrev]{cleveref}
\usepackage{multirow}
\definecolor{Gray}{gray}{0.925}
\definecolor{Green}{rgb}{0.96,1,0.96}

\usepackage{microtype}
\usepackage{graphicx}
\usepackage{subfigure}
\usepackage{booktabs} % for professional tables

\usepackage{hyperref}

\usepackage[accepted]{icml2024}

\usepackage{amsmath}
\usepackage{amssymb}
\usepackage{mathtools}
\usepackage{amsthm}

\usepackage[capitalize,noabbrev]{cleveref}

\theoremstyle{plain}

\theoremstyle{definition}

\theoremstyle{remark}

\usepackage[textsize=tiny]{todonotes}

\let\subparagraph\paragraph
\usepackage[compact]{titlesec}

\titlespacing{\section}{0pt}{1ex}{1ex}
\titlespacing{\subsection}{0pt}{1ex}{0.5ex}
\titlespacing{\subsubsection}{0pt}{0.5ex}{0.5ex}

\DeclareMathSizes{9}{8.2}{5}{3}
\icmltitlerunning{Paper in submission}

\begin{document}

\twocolumn[
\icmltitle{Question Rephrasing for Quantifying Uncertainty in Large Language Models: Applications in Molecular Chemistry Tasks}



\icmlsetsymbol{equal}{*}

\begin{icmlauthorlist}
% \icmlauthor{Anonymous Authors}{}
\icmlauthor{Zizhang Chen}{yy}
\icmlauthor{Pengyu Hong}{yy}
\icmlauthor{Sandeep Madireddy}{comp}

\icmlaffiliation{yy}{Brandeis University}
\icmlaffiliation{comp}{Argonne National Laboratory}

\end{icmlauthorlist}



\icmlkeywords{Machine Learning, ICML}

\vskip 0.3in
]




\begin{abstract}
\vspace{1.5mm}
Uncertainty quantification enables users to assess the reliability of responses generated by large language models (LLMs). We present a novel Question Rephrasing technique to evaluate the input uncertainty of LLMs, which refers to the uncertainty arising from equivalent variations of the inputs provided to LLMs. This technique is integrated with sampling methods that measure the output uncertainty of LLMs, thereby offering a more comprehensive uncertainty assessment. We validated our approach on property prediction and reaction prediction for molecular chemistry tasks.
\end{abstract}

%%%%%%%%%%%%%%%%%%%%%%%%%%%%%%%%%%%%%%%%%%%%%%%%%%%%%%%%%%%%%%%%%%%%%%%%%%

% logic: LLM good -> LLM can be applied for scientific tasks -> if LLM applied for domain tasks it requires UQ to ensure its confidence -> we choose chemistry as the domain task. 
% Dr
\section{Introduction}
\label{sec:introduction}

In recent years, Large Language Models (LLMs), such as GPT~\citep{achiam2023gpt}, Claude~\cite{anthropic2024claude}, and Llama~\cite{touvron2023llama}, have demonstrated remarkable success in various tasks. Pre-trained on vast amounts of data and boosted with billions of parameters, these LLMs demonstrated impressive capabilities across a range of scientific domains, including chemistry~\cite{guo2023indeed}, biology~\cite{agathokleous2023use}, and physics~\cite{nguyen2023astrollama}. Despite their successes, a critical aspect that remains underexplored is the uncertainty inherent in the predictions produced by these LLMs. Understanding and quantifying uncertainty in LLM outputs is crucial for several reasons. It aids in informed decision-making, enhances user trust, and ensures the safety and reliability of AI systems~\citep{sun2024trustllm}. Moreover, transparency about model uncertainty fosters responsible AI deployment. 

Inspired by the practice in psychological assessments, where clinicians ask the same question in different ways to test a patient's understanding and consistency of responses, we propose a technique, termed {\it Question Rephrasing}, to quantify the uncertainty of the answer produced by an LLM in response to a question. Essentially, given an initial question, the {\it Question Rephrasing} technique involves rephrasing the question while maximally preserving its original meaning and then submitting the rephrased question to the LLM. The consistency between the LLM's answers before and after rephrasing is evaluated to quantify the uncertainty of the LLM with respect to the input variations. In addition, a sampling approach is adopted that repeatedly queries the LLM with the same input to assess the output uncertainty of the LLM.

In our experiments, we applied our method to quantify the uncertainty of GPT-3.5/4~\citep{achiam2023gpt} on two tasks in the Chemistry domain: property prediction and forward reaction prediction analogous to classification and text generation tasks, respectively. We found that GPT-4 was sensitive to {\it Question Rephrasing}, and the output uncertainty could serve as a valuable indicator for the accuracy and reliability of the LLM's response.

%\SM{add one line on the quantitative results/conclusion}

%%%%%%%%%%%%%%%%%%%%%%%%%%%%%%%%%%%%%%%%%%%%%%%%%%%%%%%%%%%%%%%%%%%%%%%%%%

\section{Background and Related Work}
\label{sec:relate}

%smiles, IUPAC
\subsection{Textual representation of molecules}
The textual representation of molecular structures is fundamental for applying language models to chemistry-related tasks. Prominent among these representations are the Simplified Molecular Input Line Entry System (SMILES)~\citep{weininger1988smiles, o2012towards} and the International Union of Pure and Applied Chemistry (IUPAC)~\citep{panico1993guide, leigh2011principles} nomenclature. Currently, no standardized rules are in place for assigning common names to chemical compounds. IUPAC provides a universally recognized method for naming chemical entities, whereas SMILES offers a more compact, machine-readable format that has recently facilitated significant advancements in applying language models to chemistry~\citep{xu2017seq2seq, ross2022large, wu2023molformer, fang2024molinstructions}. Given its ease of use and compatibility with various machine learning workflows, we used the SMILES notation as the primary method for representing molecular structures. 
% blackbox llm, iclr 2023, jimeng's paper

\subsection{Chemistry tasks and LLMs}
% property prediction - classification tasks
% reaction prediction - generation tasks
% What have been done for the above two tasks using LLMs.

Recent literature highlights the expanding role of LLMs in molecular chemistry, particularly in enhancing predictive and generative tasks. ~\cite{guo2023can} established benchmarks for evaluating LLMs in property and reaction outcome predictions, demonstrating their broad applicability. ~\cite{zhong2024benchmarking} showed that while LLMs lag behind specialized machine learning models in processing geometric molecular data, they significantly enhance performance when integrated with these models. ~\cite{zhong2024harnessing} shows that LLMs as post-hoc correctors improves the accuracy of molecular property predictions after initial model training.
~\cite{qian2023can} and ~\cite{jablonka2024leveraging} underscore the utility of LLMs in generating explanatory content for molecular structures and resolving complex chemical queries, enhancing both educational and practical applications. ~\cite{luong2024application} found that transformer-based models like GPT and BERT exhibit high accuracy in reaction prediction and molecule generation.
% , despite challenges with structured chemical data and computational costs. Despite these limitations, the integration of LLMs with ML models shows significant promise for advancing molecular chemistry applications.

\subsection{Uncertainty quantification for black-box LLMs}

The recent shift towards black-box LLMs, particularly in commercially deployed models such as GPT4~\citep{achiam2023gpt}, Claude 3~\citep{anthropic2023claude3} and Gemini~\citep{team2023gemini}, presents unique challenges for Uncertainty Quantification (UQ). Traditionally, UQ techniques have relied heavily on accessing the internal model parameters and predictions at a granular level, such as token probabilities and logits~\citep{gal2016dropout, malinin2018predictive, hu2023uncertainty}. However, the encapsulation of modern LLMs, often provided as API services, restricts such access. Recent studies~\cite{kuhn2023semantic, lin2023generating, xiong2024can} have started to address these limitations by innovating methods and pipelines that infer uncertainty directly from the text outputs generated by LLMs without requiring their internal workings. Kuhn et al.(\citeyear{kuhn2023semantic}) introduce semantic entropy, a novel metric to quantify uncertainty in LLMs that focuses on semantic equivalence, the concept that different phrases can express the same meaning. Later works~\citep{lin2023generating, xiong2024can} introduce complex frameworks to refine black-box UQ methods comprising prompting strategies, sampling methods, and aggregation techniques. This work aims to quantify the black-box LLMs uncertainty on chemistry-related tasks. 

%%%%%%%%%%%%%%%%%%%%%%%%%%%%%%%%%%%%%%%%%%%%%%%%%%%%%%%%%%%%%%%%%%%%%%%%%%

\section{Uncertainty Quantification in Molecular Chemistry Tasks}

This section introduces and discusses UQ methods for chemistry-related tasks using black-box LLMs. We categorized our UQ metrics into two parts: \textbf{input uncertainty} and \textbf{output uncertainty}. Input uncertainty uses the {\it Question Rephrasing} strategy to assess LLM's sensitivity to variations in molecular representations. We systematically use the alternative SMILES representations of each input molecule in the prompt and investigate how these perturbations impact the LLM's output predictions. Since the alternative SMILES of the same molecule is used, we were able to guarantee that the semantics of the modified prompt remains the same. In addition, this method can test whether an LLM truly understands molecular representations in chemistry or is only able to perform string comparisons. Output uncertainty assesses the consistency of the output produced by an LLM, which is influenced purely by the model's inherent properties. We repeatedly query the model with identical input to create a distribution of the answers. 
We structured our pipelines based on existing UQ-related works~\citep{prabhakaran-etal-2019-perturbation, lin2023generating, kuhn2023semantic}. Below, we outline our UQ methods:

\begin{enumerate}
\item For a chemistry-related task $t$, given a SMILES representation $x_i$ of the $i$-th molecule, generate a prompt $P_{t, x_i}$ based on a task-specific template (see Section~\ref{sec:prompt}). % and feed it into the black-box LLM for question-answer generations.  with structure $s$.

\item  Generate a list of up to $n$ SMILES variants of the molecule $x_i$: $L=\{x_i^1, x_i^2, ..., x_i^n\}$. We ask GPT-4 to rank the SMILES variants by its confidence to interpret their structures, and choose the one, say $\hat{x}_i$, with the highest confidence to construct a prompt $P_{t, \hat{x}_i}$ by replacing $x_i$ in $P_{t, x_i}$ with $\hat{x}_i$ (see Section~\ref{sec:input_uncertainty}).

\item Ask the LLM to generate $m$ responses for the prompt $P_{t, x_i}$ and obtain $R_{t, x_i} = \{r_{t,x_i,1}, r_{t,x_i,2}, ..., r_{t,x_i,m}\}$. Ask the LLM to generate $m$ responses for the prompt $P_{t, \hat{x}_i}$ and obtain $R_{t, \hat{x}_i} = \{r_{t,\hat{x}_i,1}, r_{t,\hat{x}_i,2}, ..., r_{t,\hat{x}_i,m}\}$. 

\item Calculate the entropy-based uncertainty metrics $U_{t, x_i}$ and $U_{t, \hat{x}_i}$ for $R_{t, x_i}$ and $R_{t, \hat{x}_i}$, respectively.

\item Measure the input uncertainty by comparing $U_{t, x_i}$ and $U_{t, \hat{x}_i}$ for all chosen $x_i$. Measure the output uncertainty by examining $U_{t, x_i}$ and $U_{t, \hat{x}_i}$ separately. 
\end{enumerate}

In the subsequent subsections, we provide detailed explanations of our UQ methods.

\subsection{Prompt design for molecular chemistry tasks}
\label{sec:prompt}
It was shown that LLMs exhibited a certain degree of zero-shot learning capabilities~\citep{brown2020language}. Here, we adopted and modified the structured approach delineated in the recent Chemistry LLM benchmark study~\cite {guo2023can} to design chemistry task-specific prompt completion pairs using In-Context Learning (ICL) samples. Motivated by the OpenAI prompt guide~\citep{shieh2023best} and the benchmark paper~\cite{guo2023can}, we designed our prompts to consist of three parts: 1. Chemistry role-playing prompts with task-specific instructions. 2. Few shot ICL samples were constructed using k-scaffold sampling. 3. Questions to be answered for the target SMILES. ~\cref{tab:prompt} showcases the prompt design for the toxicity prediction task. 

\begin{table}[ht]
\centering
\caption{An example of prompts for chemistry-related tasks.}
\label{tab:prompt}
\vspace{1mm}
\begin{tabular}{|>{\raggedright\arraybackslash}p{7cm}|}
\hline
\textbf{Role}: \textit{You are an expert chemist specializing in chemical property prediction.}\\
\hline
\textbf{Task}: \textit{Given the SMILES a molecule, use your expertise to predict the molecular properties based on its structure...} \\
\hline
\textbf{ICL samples}: \textit{For the following SMILES, determine if each molecule contains a toxicity compound, answering only with "Yes" or "No". A few samples are provided:} \\
\textit{SMILES: few-shot example smiles 1} \\
\textit{Contain toxicity compound: Yes} \\
...\\
\textit{SMILES: few-shot example smiles p} \\
\textit{Contain toxicity compound: No} \\
\hline
\textbf{Question}:
\textit{SMILES: target smiles} \\
\textit{Contain toxicity compound: [Provide an answer based on analysis]} \\
\textit{Please strictly answer with "Yes" or "No".} \\
\hline
\end{tabular}
\end{table}

% change input format of the molecules
% change 
\subsection{Input Uncertainty: Sensitivity Analysis}
\label{sec:input_uncertainty}
\begin{figure}[t]
\centering
    \includegraphics[width=1\columnwidth]{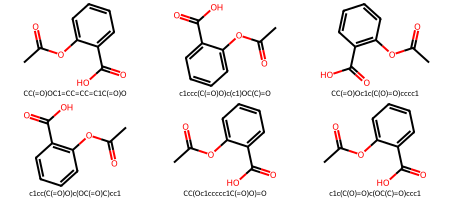}
    \vspace{-7mm}
    \caption{SMILES representation variants of Aspirin. While all structures depict the same molecule, their SMILES representations are different, which introduces input variations. \textbf{Top left}: Canonical SMILES representation of Aspirin. \textbf{Rest}: Five SMILES variations of Aspirin.}
    \label{fig:asprin}
    % \vspace{-4mm}
\end{figure}

We investigated input uncertainty by analyzing the sensitivity of a black-box LLM to changes in inputs. For each ICL prompt $P_{t,x_i}$ of a chemistry task $t$, we rephrased it by replacing the SMILES representation $x_i$ with its equivalent SMILES to generate a new prompt. Specifically, we first obtained the structure of the molecule $s_i$ of $x_i$ using RDKit~\citep{landrum2013rdkit, greg_landrum_2020_3732262}. Then, we obtained a list of up to $n$ distinct SMILES representations $L = \{x_i^1, x_i^2, ..., x_i^n\}$ for the structure $s_i$. For better illustration, we use Aspirin as an example to showcase this step (see ~\cref{fig:asprin}). We then prompted GPT-4 to rank the obtained SMILES variants by its confidence in interpreting the structures from those SMILES variants (see Table 2). The SMILES variant $\hat{x}$ with the highest confidence score was chosen to construct a new prompt $P_{t,\hat{x}_i}$ by replacing $x_i$ in $P_{t,x_i}$ with $\hat{x}_i$. The LLM was then asked to generate responses for the prompts $P_{t, x}$ and $P_{t, \hat{x}}$ separately. We then evaluated the responses produced by LLM for $P_{t, x}$ and $P_{t, \hat{x}}$. \textit{Accuracy} was the metric used in the molecule classification tasks, and \textit{exact match accuracy} was the metric used in the tasks that generate SMILES. % This metric measures the proportion of instances where the generated molecule's SMILES representation exactly matches the target SMILES. 
% data mining techniques for outputs

\begin{table}[h]
\centering
\caption{Prompt template for generating SMILE confidence score}
\label{tab:prompt2}
\vspace{1mm}
\begin{tabular}{|>{\raggedright\arraybackslash}p{7cm}|}
\hline
\textbf{Role}: \textit{As an expert in chemistry with a thorough understanding of SMILES notation. }\\
\hline
\textbf{Questions}: \textit{Can you rank your confidence score in the following smiles for interpreting its structures?  [please output the exact smile string]:} \\
\textit{variation SMILES 1} \\
\textit{variation SMILES 2} \\
...\\
\textit{variation SMILES n} \\
\hline
\end{tabular}
\end{table}

\subsection{Output uncertainty: Uncertainty Quantification from Structure Similarly}

In this section, we explain the entropy-based metrics for measuring the output uncertainty of black-box LLMs, focusing on classification and generation tasks in the chemistry domain. 
\\
For classification tasks, the LLM's responses $R_{t, x_i} = \{r_{t,x_i,1}, r_{t,x_i,2}, ..., r_{t,x_i,m}\}$ of the molecule $x_i$ can be interpreted as a set of classification results, where each response $r_{t, x_i, j}$ is a class label predicted by LLMs from a set of possible classes $C = \{c_1, c_2, \ldots, c_k\}$. Here, $k$ is the number of classes that appear in the prediction outputs. The probability of each class $c_j\in C$ can be calculated as the percentage of $c_j$ appearing in $R_{t, x_i}$: 

%\vspace{-2mm}
\begin{equation}
P(c_j) = \frac{|\{ r_{t,x_i}  = c_j : r_{t,x_i} \in R_{t, x_i} \}|}{ |R_{t, x_i}| }
\end{equation}

where $|\{ r_{t,x_i}  = c_j : r_{t,x_i} \in R_{t, x_i} \}|$ counts the number of times that class $c_j$ appears in $R_{t, x_i}$. The uncertainty score $U_{t, x_i}$ is formulated as:

%\vspace{-3mm}
\begin{equation}
\label{eq:ce}
    U_{t, x_i} = -\sum_{j=1}^k P(c_j) \log P(c_j)
\end{equation}
%\vspace{-2mm}

For all generation tasks that produce the SMILES representation, we measured the similarity between the generated SMILES using the Tanimoto Similarity~\citep{butina1999unsupervised, chung2019jaccard} based on their molecular fingerprints, which can be obtained with RDKit~\citep{landrum2013rdkit}. Sometimes an LLM may generate invalid SMILES representations. We set the similarity between an invalid SMILES and any other SMILES to be an infinitely small number $\epsilon$. Once we obtain the pairwise similarity between all SMILES generated for a specific molecule $x_i$, we applied hierarchical clustering to group the generated SMILES into $g$ clusters $S = \{s_1, s_2, \ldots, s_g\}$. The probability of a cluster $s_j\in S$ is calculated as its percentage in $R_{t, x_i}$: 

%\vspace{-2mm}
\begin{equation}
P(s_j) = \frac{|\{r_{t,x_i} \in R_{t, x_i} : r_{t,x_i}  = s_j\}|}{m}
\end{equation}
Without loss of generality, the uncertainty score $U_{t, x_i}$ can be formulated as follows:
%\vspace{-2mm}
\begin{equation}
\label{eq:se}
    U(R_{t, x_i} \mid S) = -\sum_{j=1}^g P(s_j) \log P(s_j)
\end{equation}
% We then showcase the use of uncertainty score to analysis LLM's confidences of its generated answers.
%%%%%%%%%%%%%%%%%%%%%%%%%%%%%%%%%%%%%%%%%%%%%%%%%%%%%%%%%%%%%%%%%%%%%%%%%%%%%%%%%%%

\section{Experiments}
Following~\cite{kuhn2023semantic, lin2023generating}, we evaluate our output uncertainty metric by utilizing it to predict whether LLM can correctly generate an answer. We plot the Receiver operating characteristic curve (ROC) and calculate the Area under the ROC Curve (AUC) score. An AUC score of 0.5 indicates that the uncertainty metrics are no better than a random classifier, whereas a high AUC score indicates that the metrics can help us determine whether to trust the model's response. We evaluated the input uncertainty by comparing the model performances across different inputs. A significant increase or decrease in model performance may indicate that the model is sensitive to its input and, thus, less likely to be trusted.
% what it is?
% What I do
% what are the results
% What is the conclusion
\subsection{Property Prediction}

\begin{figure}[h]
    \centering
    \includegraphics[width=1.\columnwidth]{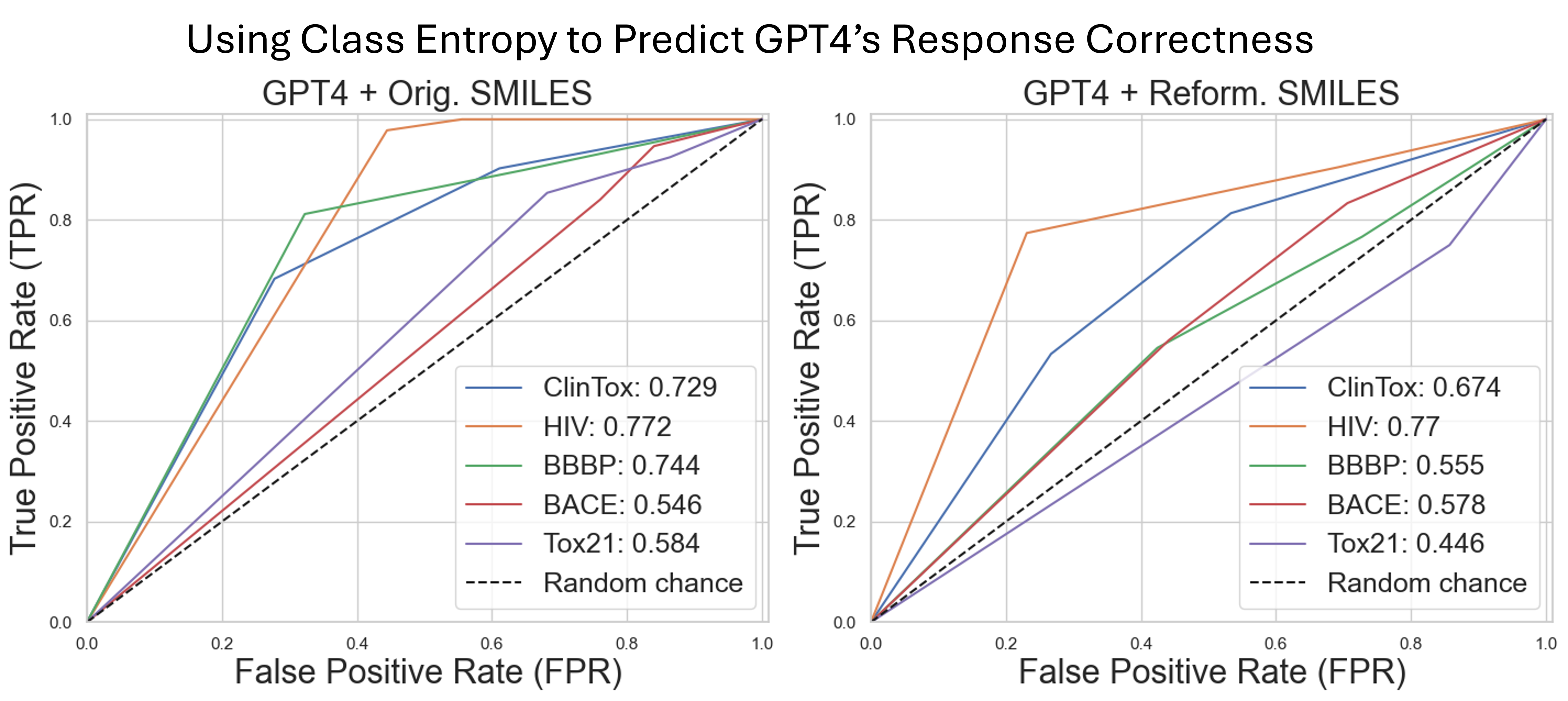}
    \vspace{-4mm}
    \caption{ROC curve for evaluating the in predicting the correctness of the GPT using our uncertainty score. }
    \label{fig:pp}
    %\vspace{-3mm}
\end{figure}

We used five datasets (BBBP, HIV, BACE, Tox21, and ClinTox~\citep{wu2018moleculenet}) and the associated tasks to investigate the capabilities of our method to quantify the uncertainty of Black-box LLMs (specifically GPT-4) on predicting molecular properties. These datasets, sourced from the corresponding established databases and scientific literature, are primarily used in training machine learning models to predict binary molecular properties from their SMILES representations. For each dataset, adapted from the experimental settings of~\citep{guo2023can}, we randomly sampled the $100$ molecules as a test set and constructed the prompts using ICL samples querying from the rest of the dataset. For each prompt, we repeatedly generated $5$ responses and calculated the uncertainty score from~\cref{eq:ce}, here, denoted as Class Entropy, and used to predict whether GPT-4 can generate the correct answers. In addition, we reformulate the input SMILES and re-run the experiments following the methods mentioned in~\cref{sec:input_uncertainty}. \\
The prediction and uncertainty quantification results are presented in~\cref{tab:pp} and ~\cref{fig:pp}. We noticed a slight decrease in model performance (except BP) when using reformed SMILES over the original SMILES input in~\cref{tab:pp}. This indicates GPT's relatively high confidence among the input invariants. In addition, according to~\cref{fig:pp}, the AUC score for the original SMILES spans between 0.546 and 0.774, indicating a moderate trustworthiness in using the output uncertainty score to predict the GPT's response correctness. 

\begin{table}[h]
\centering
\caption{Property prediction results of GPT-4 using original input SMILES (Orig. SMILES) and reformulated SMILES (Reform. SMILES) on five datasets. The evaluation metrics include Accuracy and F1 score. The average Class Entropy (C. E) is also reported.} 
\label{tab:pp}
\resizebox{1.\columnwidth}{!}{
    \begin{tabular}{l|ccc|ccc|}
    
    \midrule 
    Model & \multicolumn{3}{c}{$GPT$$-$$4$ (Orig. SMILES)} & \multicolumn{3}{c}{$GPT$$-$$4$ (Reform. SMILES)} \\
    \midrule
    Eval. metric & Acc. & F1 & C.E. & Acc. & F1 & C.E. \\
    \midrule
    BACE & 0.750  & 0.766 &  0.150 & 0.660 $\color{blue}\downarrow$ & 0.638$\color{blue}\downarrow$ & 0.398 \\

    BBBP & 0.690 & 0.756 & 0.290 & 0.700 $\color{red}\uparrow$ & 0.795 $\color{red}\uparrow$ & 0.415 \\
     
    ClinTox & 0.820 & 0.357 & 0.319 & 0.833 $\color{blue}\downarrow$   & 0.285 $\color{blue}\downarrow$   & 0.427 \\
    
    HIV & 0.910  & 0.471 & 0.060 & 0.763 $\color{blue}\downarrow$   & 0.350 $\color{blue}\downarrow$   & 0.292 \\
     
    Tox21 & 0.707 & 0.522 & 0.105 & 0.533 $\color{blue}\downarrow$   & 0.416 $\color{blue}\downarrow$ & 0.290 \\
    \bottomrule
    \end{tabular}
}
\end{table}

%\vspace{-4mm}

%%%%%%%%%%%%%%%%%%%%%%%%%%%%%%%%%%%%%%%%%%%%%%%%%%%%%%%%%%%%%%%%%%%%%%%%%%%%%%%%%%%

\subsection{Forward Reaction Prediction}

We utilize the USPTO-MIT dataset~\citep{schneider2016s, jin2017predicting} to evaluate our uncertainty quantification metrics. The test set is constructed by randomly sampling 100 reaction-product pairs, while the remaining data are used to query the in-context learning (ICL) samples. For evaluations, we employ GPT-4 and GPT-3.5 Turbo to generate responses. We repeatedly generate 3, 10, 15, and 20 responses for each prompt. We first calculate the accuracy score by performing an exact match comparison between the generated SMILES and the ground-truth SMILES. We then calculate the output uncertainty metric and use it to predict whether the response from black-box LLMs is correct. We then derived the AUC score for each set of responses. In addition, we perform the input uncertainty analysis by reformulating the input SMILES as we mentioned in~\cref{sec:input_uncertainty} and repeat the above steps. \\
We present our results in~\cref{tab:react}. We observe that GPT-3.5/4 performed poorly on reaction prediction tasks. In addition, our output uncertainty metrics are reliable indicators of the correctness of GPT's responses (AUC score ranges from 0.86 to 0.99). We also observed a substantial decline in model performance on reaction prediction tasks when presented with the variations in molecular representation, demonstrating the LLMs' weakness in understanding basic chemistry knowledge. 

\begin{table}[h]
\centering
\vspace{-1mm}
\caption{Reaction prediction performances of GPTs and AUC scores of output uncertainty metrics} 
\label{tab:react}
\resizebox{1.\columnwidth}{!}{
    \begin{tabular}{l|c|ccccc}
    \midrule 
    Method & Top-1 Acc. & AUC-3 & AUC-10 & AUC-15 & AUC-20\\
    % \hline Chemformer & 0.938 & $0 \%$ \\
    \midrule
    GPT-4 + Orig.& 0.250  & 0.864 & 0.919 & 0.915 & 0.927\\
    GPT-4 + Reform & 0.070 $\color{blue}\downarrow$ & 0.972 & 0.941 & 0.958 & 0.993\\
    \midrule
    GPT-3.5 + Orig & 0.186  &  0.904 & 0.899 &  0.924 & 0.943\\
    GPT-3.5 + Reform & 0.036 $\color{blue}\downarrow$ & 0.919 & 1.000 & 1.000 & 1.000 \\
    \bottomrule
    \end{tabular}
}
\end{table}

\section{Conclusions}
% We proposed 

% We experimentally showcased the output uncertainty metric effectively predicted the correctness and reliability of the LLM's responses in chemistry tasks, and the input uncertainty can be used as an assistant 

% use as needed
In this work, we introduce a novel {\it Question Rephrasing} technique for uncertainty quantification in LLMs, specifically applied to chemistry tasks. By integrating input and output uncertainty assessments, we enhanced the ability to comprehensively evaluate the reliability of LLMs. We applied our approach to quantify the trustworthiness of LLMs in molecular chemistry. Experiment results show that GPT-3.5/4 exhibits sensitivity to input variations, and entropy-based metrics can effectively capture the output uncertainty of GPT-3.5/4, enabling the prediction of the correctness of LLM responses. Our experimental results underscore the need to enhance LLMs' understanding of basic chemistry knowledge. We believe that our approach and the discovery in this study help pave the way for developing more reliable and transparent AI systems for scientific applications.


\newpage
\bibliography{example_paper}
\bibliographystyle{icml2024}

%%%%%%%%%%%%%%%%%%%%%%%%%%%%%%%%%%%%%%%%%%%%%%%%%%%%%%%%%%%%%%%%%%%%%%%%%%%%%%%
%%%%%%%%%%%%%%%%%%%%%%%%%%%%%%%%%%%%%%%%%%%%%%%%%%%%%%%%%%%%%%%%%%%%%%%%%%%%%%%
% APPENDIX
%%%%%%%%%%%%%%%%%%%%%%%%%%%%%%%%%%%%%%%%%%%%%%%%%%%%%%%%%%%%%%%%%%%%%%%%%%%%%%%
%%%%%%%%%%%%%%%%%%%%%%%%%%%%%%%%%%%%%%%%%%%%%%%%%%%%%%%%%%%%%%%%%%%%%%%%%%%%%%%
\newpage
\appendix
\onecolumn
%%%%%%%%%%%%%%%%%%%%%%%%%%%%%%%%%%%%%%%%%%%%%%%%%%%%%%%%%%%%%%%%%%%%%%%%%%%%%%%
%%%%%%%%%%%%%%%%%%%%%%%%%%%%%%%%%%%%%%%%%%%%%%%%%%%%%%%%%%%%%%%%%%%%%%%%%%%%%%%

\end{document}